\documentclass{article}
\usepackage{spconf,amsmath,graphicx}

\usepackage{color}
\usepackage{xcolor}
\usepackage{comment}
\usepackage{indentfirst}
\usepackage{algorithm}
\usepackage{algpseudocode}
\usepackage{bbm}
\usepackage{booktabs}
\usepackage{soul}
\usepackage{lipsum}
\usepackage{hyperref}
\usepackage{cleveref}
\usepackage{amssymb}

\title{Contextuality Helps Representation Learning for Generalized Category Discovery}
\name{Tingzhang Luo $^{1}$, Mingxuan Du$^1$, Jiatao Shi$^1$,Xinxiang Chen$^1$, Bingchen Zhao$^{2}$ and Shaoguang Huang$^1$\thanks{This work is supported in part by the ``CUG Scholar'' Scientific Research Funds at China University of Geosciences (Wuhan) under grant 2022164, in part by the National Natural Science Foundation of China under grant 42301425, and in part by the China Postdoctoral Science Foundation (2023M743299).}}
\address{
	$^1$ China University of Geosciences, China \\
	$^2$ University of Edinburgh, United Kingdom
}

\begin{document}
	%
	\maketitle
	%
	\begin{abstract}
		This paper introduces a novel approach to Generalized Category Discovery (GCD) by leveraging the concept of contextuality to enhance the identification and classification of categories in unlabeled datasets. Drawing inspiration from human cognition's ability to recognize objects within their context, we propose a dual-context based method. 
		Our model integrates two levels of contextuality: instance-level, where nearest-neighbor contexts are utilized for contrastive learning, and cluster-level, employing prototypical contrastive learning based on category prototypes. The integration of the contextual information effectively improves the feature learning and thereby the classification accuracy of all categories, which better deals with the real-world datasets. Different from the traditional semi-supervised and novel category discovery techniques, our model focuses on a more realistic and challenging scenario where both known and novel categories are present in the unlabeled data.  Extensive experimental results on several benchmark data sets demonstrate that the proposed model outperforms the state-of-the-art. Code is available at: \url{https://github.com/Clarence-CV/Contexuality-GCD}
	\end{abstract}
	\begin{keywords}
		Generalized category discovery, semi-supervised learning, contrastive learning, clustering
	\end{keywords}
	\section{Introduction}
	\label{sec:intro}

	Deep learning techniques have significantly boosted the performance in computer vision tasks. In the field of image classification, the performance using deep learning already approached or even beyond human-level proficiency~\cite{he2015delving}. However, these advanced techniques are often difficult to be deployed in real applications due to the lack of labeled data. Existing techniques to address the problem of scarce labeled data include deep clustering and semi-supervised learning ~\cite{lee2013pseudo,zhang2017mixup,sohn2020}, where the former does not use labeled data and the latter uses both labeled data and unlabeled data. Although the semi-supervised approaches often obtain superior performance than clustering methods, they are limited by the implicit assumption that the classes in the labeled data and unlabeled data are the same, which is often not true in real applications.
	
		\begin{figure}
		\centering
		\includegraphics[width=.8\linewidth]{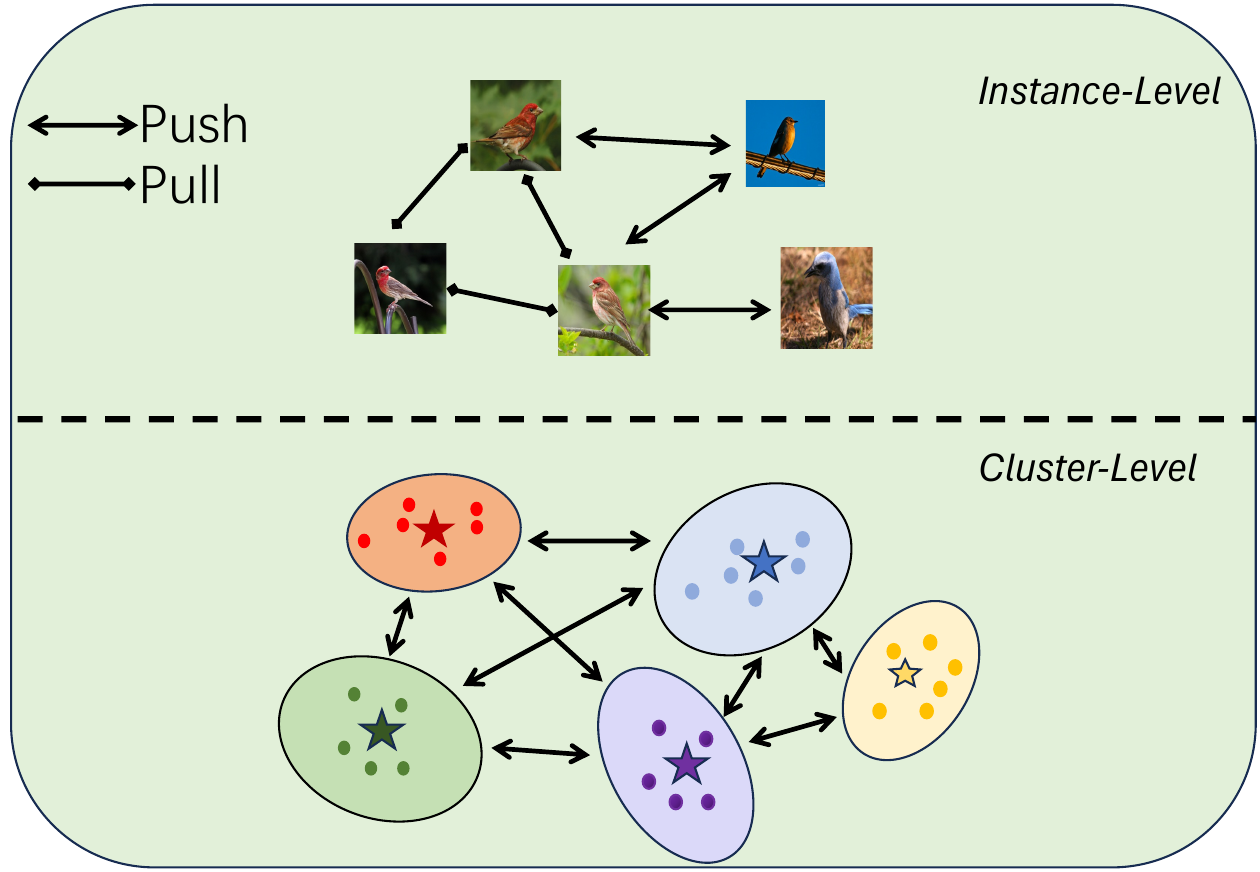}
		\caption{
			Two levels of contextuality are explored in our method. 
			The instance-level context leverages the nearest-neighbor context and pseudo-labels to search pair-wise data points for contrastive learning. The cluster-level context forms prototypes to learn the representation via prototypical contrastive learning.
		}
		\label{fig:teaser}
		\vspace{-.5cm}
	\end{figure}
	
	Recently, Generalized Category Discovery (GCD)~\cite{vaze2022generalized} is formalized to relax the assumption by assuming that the unlabeled data contains the categories of labeled data and also novel unseen categories. The goal of GCD is to classify all the categories in the unlabeled data based on existing knowledge from a set of already labeled categories. Generally, GCD is more challenging compared with the setting in semi-supervised learning as there are no labeled training data for the novel categories in the unlabeled data. Some attempts have been made towards GCD tasks, which often involve feature representation and classifier learning \cite{wen2023parametric}. For instance, the work in \cite{vaze2022generalized} adopts contrastive learning technique to improve the generalization ability of features to novel categories and uses a non-parametric classifier k-means to obtain final clustering results. In \cite{wen2023parametric,cao2021orca,fini2021unified,han2021autonovel}, parametric approaches are developed for a learnable classifier by using both labeled and pseudo-labeled data. The incremental setting of GCD is also investigated in \cite{zhao2023incremental}. These methods have obtained remarkable performance improvement in GCD tasks. However, they often focus on instance-level supervision information and overlook the underlying group relationships between samples or categories, which might lead to degraded representation learning. Neuro-scientific researchers suggest that human cognition excels in recognition tasks through contextual understanding~\cite{palmer1975effects}. Humans recognize and categorize objects effectively by considering their context, not just their isolated features.
	
	Therefore, we identify two distinct levels of context as shown in Fig. \ref{fig:teaser} to enhance the feature learning in GCD:
	
	\begin{itemize}
		\item \textbf{Instance-Level Nearest-Neighbor Context}: This involves leveraging nearest-neighbor contexts to generate pairwise labels for contrastive learning, encouraging the model to group examples with similar contexts and separate those with dissimilar contexts, akin to human cognitive processes.
		\item \textbf{Cluster-Level Context}: Here, we average the representations of each discovered category to form a prototype, representing an abstract concept of the category. Prototypical contrastive learning is then applied to these cluster prototypes across views, enhancing the discriminative power of the representations.
	\end{itemize}
	
	By incorporating the contextual information at different levels for feature learning, we propose a novel model for GCD. The main contributions of this paper can be summarized as follows:
	\begin{enumerate}
		\item We propose a novel contextual information guided semi-supervised learning model to effectively leverage the underlying relationships between instances at multiple levels, which facilitates learning discriminative features of novel categories in GCD.
		\item We develop a self-supervised contrastive learning strategy by incorporating neighborhood information and pseudo-label supervision information at instance level to reduce the within-cluster variance. We also consider the group effect of instances and develop a cluster-level contrastive loss across views for feature learning.
		\item We conduct extensive experiments on benchmark data sets and demonstrate the superior performance of our model over the state-of-the-art.
	\end{enumerate}

	\section{Related work}

	
	\noindent
	\textbf{Semi-Supervised Learning (SSL)} makes decision by making use of both labeled and unlabeled data, which is important when labeled training data is scarce. 
	In the conventional closed-set SSL paradigm, it is assumed that both the labeled and unlabeled datasets have identical categories, with each category represented in the labeled subset. 
	Techniques like pseudo-labeling and consistency regularization, well-established in this domain, yield improved performance in the field~\cite{lee2013pseudo,zhang2017mixup}. 
	Recent innovations such as FixMatch~\cite{sohn2020} and FlexMatch~\cite{zhang2021flexmatch} have introduced confidence-based thresholding to enhance the reliability of pseudo labels.
	Despite these advances, the inherent limitation of the closed-set assumption in traditional SSL still remains, which makes it less applicable to scenarios like GCD. 
	Although some recent studies have ventured into open-set SSL by relaxing the strict equivalence of classes between labeled and unlabeled data~\cite{saito2021openmatch,yu2020multi}, their primary objective remains the enhancement of performance within already labeled categories. 
	As a result, these open-set SSL approaches are not suitable for the more complex and realistic challenges posed by category discovery tasks.

	
	\noindent
	\textbf{Category Discovery} aims to find novel categories in unlabeled data.  
	In addition to unlabeled data, labeled data is also provided at training time which gives information on the types of visual concepts that are to be discovered. 
	Novel category discovery (NCD) assumes that the classes are not overlapped in labeled dataset and unlabeled dataset~\cite{han2019learning,han2021autonovel,zhao2021novel,fini2021unified}. 
	However, in a more realistic GCD setting, unlabeled dataset contains the categories in the labeled dataset and also the novel categories~\cite{vaze2022generalized,wen2023parametric,pu2023dynamic,zhang2023promptcal}. 
	Previous works develop effective methods for the category discovery task by leveraging parametric classifiers~\cite{wen2023parametric}, improving representation learning~\cite{pu2023dynamic,fei2022xcon}, or leveraging prompting learning with larger models~\cite{zhang2023promptcal}.

	\noindent
	\textbf{Contrastive Learning} has become an effective tool for representation learning~\cite{chen2020b}. 
	This approach typically involves aligning positive examples closer together while distancing negative examples. 
	Incorporation of labeled information into contrastive learning frameworks has been explored to enhance their efficacy~\cite{khosla2020}. 
	A representative work is prototypical contrastive learning~\cite{li2021b}, which utilizes dataset-wide $k$-means clustering to generate prototypes to extract high-level features. 
	However, the process of assigning pseudo-labels once per epoch in this method can be computationally expensive.
	IDFD~\cite{tao2021clustering} enhances prototypical contrastive learning through feature decorrelation. In~\cite{huang2022learning}, ProPos introduces a prototype scattering loss that dynamically computes prototypes for each training mini-batch based on pseudo-labels assigned per epoch, further improving representation learning. 


	\section{Methodology}
	
	In this section, we first give the problem setting of GCD. Then, a baseline model SimGCD is introduced. Finally, we propose an improved model of the baseline by leveraging contextual information at instance-level and cluster-level.
	
	\subsection{Problem Setting}
	
	In GCD, we denote the unlabeled dataset by $\mathcal{D}^u=\left\{(\mathbf{x}_i^u, \mathbf{\hat{y}}_i^u)\right\}\in \mathcal{X}\times \mathcal{Y}_u$, where $\mathcal{Y}_u$ is the label space of the unlabeled data points. The objective of GCD is to learn a model that can effectively categorize the data points in $\mathcal{D}^u$ using knowledge from a labeled dataset $\mathcal{D}^l=\left\{(\mathbf{x}_i^l, \mathbf{y}_i^l)\right\}\in \mathcal{X}\times \mathcal{Y}_l$, where $\mathcal{Y}_l$ is the label space of the labeled data points and $\mathcal{Y}_l \subset \mathcal{Y}_u$.
	The number of categories in the unlabeled space, $\mathcal{Y}_u$, is represented by $K_u=|\mathcal{Y}_u|$.

	\subsection{Baseline for GCD}
	Our model is built on the framework of SimGCD~\cite{wen2023parametric}, which includes two key components, i.e., representation learning to learn discriminative representations and classifier learning for label assignment.
	
	\subsubsection{Representation Learning}
	\label{sec:simgcd}
	
	The objective of representation learning is to obtain discriminative features, which allows the classifier effectively classify all the categories.
	To do so, a pre-trained feature extractor $f(\cdot)$ is fine-tuned with two contrastive losses. Given two random augmentations $\hat{\mathbf{x}}_i$ and $\tilde{\mathbf{x}}_i$ of an image $\mathbf{x}_i$ in a training mini-batch $B$ that contains both labeled and unlabeled data, the self-supervised contrastive loss is formulated as: 
	\begin{equation}
		\mathcal{L}_{\text{rep}}^{u}=\frac{1}{|B|} \sum_{\mathbf{x}_i\in B} - \log \frac{\exp (\hat{\mathbf{z}}_i^\top \tilde{\mathbf{z}}_i / \tau_u)}{\sum_{\mathbf{x}_j \in B} \exp (\hat{\mathbf{z}}_j^\top \tilde{\mathbf{z}}_i / \tau_u)},
	\end{equation}
	where $\mathbf{z}_i=g(f(\mathbf{x}_i))$ is the projected feature for contrastive learning, $g(.)$ is the projection head, and $\tau_u$ is a temperature value.
	A supervised contrastive loss is similarly defined by:
	\begin{equation}
		\mathcal{L}_{\text{rep}}^{s} = \frac{1}{|B^l|}\sum_{\mathbf{x}_i\in B^l}\frac{1}{|\mathcal{N}_i|}\sum_{p\in\mathcal{N}_i}-\log\frac{\exp (\hat{\mathbf{z}}_i^\top \tilde{\mathbf{z}}_p / \tau_s)}{\sum_{n\neq i}\exp (\hat{\mathbf{z}}_i^\top \tilde{\mathbf{z}}_n / \tau_s)},
	\end{equation}
	where $\mathcal{N}_i$ is the indexes of images in the mini-batch that have the same label to $\mathbf{x}_i$, $B^l$ is the mini-batch of labeled training data and $\tau_s$ is a temperature value. The two losses are combined to learn the representation: $\mathcal{L}_{\text{rep}}=(1-\lambda) \mathcal{L}_{\text{rep}}^{u} + \lambda \mathcal{L}_{\text{rep}}^{s}$, where $\lambda$ is a hyperparameter.
	
	\subsubsection{Classifier Learning}
	A parametric classifier learning module is desgined to perform classification with the input of the features obtained by $f(\cdot)$.
	It is assumed that the number of categories $K=|\mathcal{Y}_u|$ is given as in~\cite{wen2023parametric,fei2022xcon}. A set of parametric prototypes for each category $\mathcal{T}=\{\mathbf{t}_1, \mathbf{t}_2, \dots, \mathbf{t}_K\}$ is randomly initialized at beginning. During the training, the soft label $\hat{\mathbf{p}}_i^k$ for each augmented view $\mathbf{x}_i$ is calculated by using a softmax function on the cosine similarity between the hidden feature and the prototypes:
	
	\begin{equation}
		\hat{\mathbf{p}}_i^k = \frac{\exp\left(\frac{1}{\tau_s}({\hat{\mathbf{h}}_i} / {\|\hat{\mathbf{h}}_i\|_2})^\top ({\mathbf{t}_k}/{\|\mathbf{t}_k\|_2})\right)}{\sum_{j}\exp{\left(\frac{1}{\tau_s} (\hat{\mathbf{h}}_i/\|\hat{\mathbf{h}}_i\|_2)^\top (\mathbf{t}_j/\|\mathbf{t}_j\|_2)\right)}},
	\end{equation}
	where $\hat{\mathbf{h}}_i=f(\hat{\mathbf{x}}_i)$ is the representation of $\hat{\mathbf{x}}_i$. Similarly, one can obtain the soft label $\tilde{\mathbf{p}}_i$ of the view $\tilde{\mathbf{x}}_i$. Then, the supervised and unsupervised losses of the classifier are formulated by:
	\begin{align}
		\mathcal{L}_{\text{cls}}^l &= \frac{1}{|B^l|}\sum_{\mathbf{x}_i\in B^l} \mathcal{L}_{\text{ce}}(\mathbf{y}_i, \hat{\mathbf{p}}_i), \\
		\mathcal{L}_{\text{cls}}^u &= \frac{1}{|B|}\sum_{\mathbf{x}_i\in B} \mathcal{L}_{\text{ce}}(\tilde{\mathbf{p}}_i, \hat{\mathbf{p}}_i) - \epsilon H(\overline{\mathbf{p}}),
	\end{align}
	where $\mathbf{y}_i$ is the ground truth label for the labeled data point $\mathbf{x}_i$, $\mathcal{L}_{\text{ce}}$ is the cross-entropy loss, and $H(\overline{\mathbf{p}})=-\sum \overline{\mathbf{p}} \log \overline{\mathbf{p}}$ regularizes the mean prediction $\overline{\mathbf{p}}=\frac{1}{2|B|}\sum_{\mathbf{x}_i\in B}(\hat{\mathbf{p}}_i + \tilde{\mathbf{p}}_i)$ in a mini-batch.
	The final classifier loss is given by $\mathcal{L}_{\text{cls}}=(1-\lambda) \mathcal{L}_{\text{cls}}^u+\lambda \mathcal{L}_{\text{cls}}^l$. 
	
	Combining the losses of the representation learning and the classifier learning, the overall loss of the baseline SimGCD is formulated by:
	\begin{equation}
		\mathcal{L}_{\text{baseline}} = \mathcal{L}_{\text{rep}} + \mathcal{L}_{\text{cls}}.
	\end{equation}

	\subsection{Contextuality Mining}
	The aforementioned baseline only utilizes the instance-level supervised information in the feature learning, which overlooks the contextual information between samples or categories. Such contextual information can be important to recognize the novel categories in the unlabeled data. Incorporating contextual information into the feature learning also means a joint decision by aggregating information from multiple data points, leading to a more robust performance than existing methods. To this end, we propose two strategies to mine the contextual information at instance-level and cluster-level, resulting in two novel contextuality-based losses for feature representation.

	\noindent \textbf{Neighborhood Contextual Mining.} 
	We first consider the contextual similarity at instance-level based on local neighborhoods and the pseudo-label information. The idea is that if two data points appear within the top-$k$ nearest neighbors of each other and share the same pseudo label, they are viewed as a contextually similar pair, which are expected to yield similar representations. We define the $k$-reciprocal nearest neighbors of a data point $\mathbf{x}_i$ in the embedding space as follows:
	\begin{equation}
		R_k(\mathbf{z}_i) = \{\mathbf{z}_j | \mathbf{z}_j \in N_{k}(\mathbf{z}_i) \text{ and } \mathbf{z}_i \in N_k(\mathbf{z}_j)\},
	\end{equation}
	where $N_k(\mathbf{z}_i)$ denotes the set of $k$-nearest neighbors of $\mathbf{x}_i$. Consequently, the contextually similar data points $\mathbf{x}_j$ to $\mathbf{x}_i$ can be obtained by:
	\begin{equation}
		C_k(\mathbf{z}_i) = \{\mathbf{z}_j | \mathbf{z}_j \in R_k(\mathbf{z}_i) \text{ and } \mathbf{p'}_i = \mathbf{p'}_j\},
	\end{equation}
	where $\mathbf{p'}_i$ and $\mathbf{p'}_j$ represent the pseudo labels of $\mathbf{x}_i$ and $\mathbf{x}_j$, respectively.
	With the contextual similar data points, we formulate our instance-level contextual contrastive loss by
	\begin{equation}
		\label{eq:ln}
		\mathcal{L}_\text{n} = \frac{1}{|B|^2 - |B|} \sum_{\mathbf{x}_i \in B, \mathbf{x}_j \in B, i\neq j}s_{ij} d_{ij} + (1 - s_{ij}) (\delta - d_{ij}),
	\end{equation}
	where $s_{ij}$ represents the similarity between $\mathbf{x}_i$ and $\mathbf{x}_j$ in the feature space, $s_{ij}=1$ if $\mathbf{z}_j$ is a contextually similar data point to $\mathbf{z}_i$, i.e., $\mathbf{z}_j \in C_k(\mathbf{z}_i)$ and $s_{ij}=0$ otherwise, $d_{ij}$ is the cosine distance between $\mathbf{z}_i$ and $\mathbf{z}_j$ and $\delta$ is a margin.

	The neighborhood contextual mining approach identifies semantic similar data points in the feature space by checking if they are contextually similar. With Eq. (\ref{eq:ln}), the contextually similar pairs are promoted to be pulled together to form a more compact representation space and the dissimilar pairs are pushed away to be more separable in the representation space.

	\noindent \textbf{Cluster Contextual Mining.}
	While neighborhood contextual mining effectively captures instance-level similarity, it offers limited global context of the whole dataset, i.e., the cluster effect of data points. Thus, we propose a cluster-level contextual mining approach across different views by regularizing the prototypes of each class across the views to be similar. The loss function is formulated by:
	\begin{equation}
		\mathcal{L}_\text{c}=\frac{1}{K}\sum_{i=1}^{K}- \log \frac{\exp(\hat{\mathbf{\mu}}_i ^\top \tilde{\mathbf{\mu}}_i / \tau)}{\sum_{j}\exp(\hat{\mathbf{\mu}}_i ^\top \tilde{\mathbf{\mu}}_j / \tau)},
	\end{equation}
	where $\hat{\mathbf{\mu}}_i$ and $\tilde{\mathbf{\mu}}_i$ represent the prototypes of the $i$-th class in two views obtained by distinct augmentations, respectively. The prototypes can be obtained by the mean of features within each class in the feature space:
	
	\begin{equation}
		\mathbf{\mu}_k = \frac{\sum_{\mathbf{z}_i\in G_{k}} \mathbf{z}_i}{\|\sum_{\mathbf{z}_i\in G_{k}} \mathbf{z}_i\|_2},
	\end{equation}
	where $G_{k}=\{\mathbf{z}_i | \mathbf{p'}_i = k\}$ denotes the set of representations $\mathbf{z}_i$ pseudo-labeled as category $k$.
	This process of prototype computation aggregates information at the cluster level, which is then utilized for prototypical contrastive learning to guide the model to obtain view-consistent representations.
	
	The prototypical contrastive loss at cluster-level facilitates to align the prototypes of the same category across the two views and maximize the distances between different categories, thereby improving the classification performance of the classifier.
	
	\noindent \textbf{Batch Construction}
	The composition of training mini-batches plays a pivotal role in the representation learning process, as evidenced by recent studies~\cite{kim2022self,roth2020revisiting}. Here, we give special attention to the construction of training mini-batches, adapting the methodology from~\cite{kim2022self} with a minor modification to suit our context mining techniques.
	For each training mini-batch, we begin by randomly selecting $q$ query instances. For each query, the top $k-1$ nearest neighbors are retrieved from the dataset, forming a mini-batch of $qk$ instances. This method ensures that each instance in the mini-batch is accompanied by a sufficient contextual backdrop, which is essential for our proposed context mining techniques. In addition, we randomly sample $M$ additional data points from the dataset. These data points are not specifically tied to the context of the $qk$ instances but are included to support the learning of the original $\mathcal{L}_\text{baseline}$ loss. Consequently, each mini-batch comprises $qk+M$ data points, balancing both the need for contextual relevance and the requirement of the $\mathcal{L}_\text{baseline}$ loss in the training process.

	\subsection{Overall Loss} 
	By integrating our bi-level contextual losses into the framework of the baseline in Section \ref{sec:simgcd}, the loss function of our proposed model for GCD is obtained by:
	\begin{equation}
		\mathcal{L} = \mathcal{L}_\text{baseline} + \lambda_{n} \mathcal{L}_{n} + \lambda_{c} \mathcal{L}_{c},
	\end{equation}
	where $\lambda_{n}$ and $\lambda_{c}$ are two balancing parameters.
	The proposed model will first be trained with $T_\text{warm}$ epochs with only the $\mathcal{L}_\text{baseline}$ loss for warmup. After the warmup, the obtained model is capable to recognize different categories, which is further fine-tuned by the overall loss.
	
		\begin{table*}[t]
		\vspace{-.5cm}
		\caption{Results on SSB in terms of ACC}
		\centering
			\begin{tabular}{@{}cccccccccc@{}}
				\toprule
				& \multicolumn{3}{c}{CUB} & \multicolumn{3}{c}{Stanford Cars} & \multicolumn{3}{c}{FGVC-Aircraft} \\ \cmidrule(l){2-4} \cmidrule(l){5-7} \cmidrule{8-10}
				Methods       & All & Old & New  & All  & Old & New & All & Old & New \\ 
				\midrule
				$k$-means~\cite{macqueen1967some_kmeans} & 34.3 & 38.9 & 32.1 & 12.8 & 10.6 & 13.8 & 16.0 & 14.4 & 16.8 \\
				RS+~\cite{han2021autonovel}          & 33.3 & 51.6 & 24.2 & 28.3 & 61.8 & 12.1 & 26.9 & 36.4 & 22.2 \\
				UNO+~\cite{fini2021unified}               & 35.1 & 49.0 & 28.1 & 35.5 & 70.5 & 18.6 & 40.3 & 56.4 & 32.2 \\
				ORCA~\cite{cao2021orca}                     & 35.3 & 45.6 & 30.2 & 23.5 & 50.1 & 10.7 & 22.0 & 31.8 & 17.1 \\
				\midrule
				GCD~\cite{vaze2022generalized}       & {51.3} & {56.6} & {48.7} & {39.0} & 57.6 & {29.9} & {45.0} & 41.1 & {46.9} \\
				XCon~\cite{fei2022xcon}   & 52.1 & 54.3 & 51.0 & 40.5 & 58.8 & 31.7 & 47.7 & 44.4 & 49.4 \\
				PromptCAL~\cite{zhang2023promptcal} & 62.9 & 64.4 & \ul{62.1} & 50.2 & 70.1 & 40.6 & 52.2 & 52.2 & \ul{52.3} \\
				DCCL~\cite{pu2023dynamic} & \ul{63.5} & 60.8 & \textbf{64.9} & 43.1 & 55.7 & 36.2 & -    &  -   & - \\
				SimGCD~\cite{wen2023parametric}     & 60.3 & \ul{65.6} & 57.7 & \ul{53.8} & \ul{71.9} & \ul{45.0} & \ul{54.2} & \ul{59.1} & 51.8 \\ 
				Ours                      & \textbf{64.5} & \textbf{68.7} & 61.5 & \textbf{55.8} &\textbf{73.1} & \textbf{47.8} & \textbf{56.8} & \textbf{61.8} & \textbf{54.4}  \\
				
				\bottomrule
			\end{tabular}
		\label{tab:ssb}
		\vspace{-.5cm}
	\end{table*}
	
	\begin{table*}[t]
		\caption{Results on  generic image recognition datasets in terms of ACC}
		\centering
			\begin{tabular}{@{}cccccccccc@{}}
				\toprule
				& \multicolumn{3}{c}{CIFAR10} & \multicolumn{3}{c}{CIFAR100} & \multicolumn{3}{c}{ImageNet-100} \\ \cmidrule(l){2-4} \cmidrule(l){5-7} \cmidrule{8-10}
				Methods       & All & Old & New  & All  & Old & New & All & Old & New \\ \midrule
				$k$-means~\cite{macqueen1967some_kmeans} & 83.6 & 85.7 & 82.5 & 52.0 & 52.2 & 50.8 & 72.7 & 75.5 & 71.3 \\
				RS+~\cite{han2021autonovel}          & 46.8 & 19.2 & 60.5 & 58.2 & {77.6} & 19.3 & 37.1 & 61.6 & 24.8 \\
				UNO+~\cite{fini2021unified}               & 68.6 & \textbf{98.3} & 53.8 & 69.5 & 80.6 & 47.2 & 70.3 & \textbf{95.0} & 57.9 \\
				ORCA~\cite{cao2021orca}                     & 81.8 & 86.2 & 79.6 & 69.0 & 77.4 & 52.0 & 73.5 & {92.6} & 63.9 \\
				\midrule
				GCD~\cite{vaze2022generalized}       & {91.5} & \ul{97.9} & {88.2} & {73.0} & 76.2 & {66.5} & {74.1} & 89.8 & 66.3 \\
				XCon~\cite{fei2022xcon}   & 96.0 & 97.3 & 95.4 & 74.2 & {81.2} & 60.3 & 77.6 & 93.5 & 69.7    \\
				PromptCAL~\cite{zhang2023promptcal} & \textbf{97.9} &96.6& \ul{98.5} & \ul{81.2}& \textbf{84.2}& 75.3 & \ul{83.1}& 92.7& \ul{78.3} \\
				DCCL~\cite{pu2023dynamic} & 96.3& 96.5& 96.9& 75.3& 76.8 &70.2& 80.5& 90.5& 76.2 \\
				SimGCD~\cite{wen2023parametric}     & 97.1 & 95.1 & 98.1 & 80.1 & 81.2 & \ul{77.8} & 83.0 & 93.1 & 77.9 \\ 
				Ours                      & \ul{97.3} & 96.1 & \textbf{98.8} & \textbf{82.0} & \ul{83.9} & \textbf{78.9} & \textbf{84.6} & \ul{94.0} & \textbf{81.0} \\
				
				\bottomrule
			\end{tabular}
		\label{tab:generic}
	\end{table*}

	\section{Experiments}
	\subsection{Experimental Setup}
	\noindent
	\textbf{Datasets}. We validate the performance of our model on eight benchmark data sets, including the recently proposed Semantic Shift Benchmark (SSB) data sets CUB~\cite{welinder2010}, Stanford Cars~\cite{krause2013} and FGVC-Aircraft~\cite{maji2013},  two general image recognition data sets CIFAR10/100~\cite{krizhevsky2009} and ImageNet-100~\cite{tian2020a}, and the harder data sets Herbarium 19~\cite{tan2019herbarium} and ImageNet-1k~\cite{deng2009imagenet}. For each dataset, we follow GCD~\cite{vaze2022generalized} and take a subset of all categories as the seen (`Old') category $\mathcal{Y}_l$; 50\% of the images in these labeled categories are used for constructing $\mathcal{D}^l$, and the rest of the images are considered unlabeled data $\mathcal{D}^u$.

	\noindent
	\textbf{Evaluation protocol}. We use clustering accuracy (ACC) to evaluate the performance of different models as \cite{vaze2022generalized}.
	Given the ground truth $y^*$ and the predicted labels $\hat{y}$, the ACC is calculated as $\text{ACC} = \frac{1}{M} \sum^{M}_{i=1} \mathbbm{1}(y^{*}_{i} = p(\hat{y}_i))$, where $M = |\mathcal{D}^u|$, and $p$ is the optimal permutation that aligns the predicted cluster assignments with the ground truth class labels.
	We report the accuracy of the `All', `Old', and `New' categories, which stand for the accuracies on all categories, seen categories, and novel categories, respectively.
	
	\noindent
	\textbf{Implementation details}.
	We employ a ViT-B/16 backbone network~\cite{dosovitskiy2021} pre-trained with DINO~\cite{caron2021}, following the same setting of~\cite{vaze2022generalized}.
	The output of [CLS] token with a dimension of 768 is used as the image feature, i.e., $f(\cdot)$. Only the last block of the backbone is fine-tuned. We set the batch size to 128 and the number of epochs to 200. The learning rate is initialized to 0.1, which is decayed with a cosine schedule on each data set. We set the balance factor $\lambda$ to 0.35, and the temperature values $\tau_c$ and $\tau_u$ to 0.1 and 0.07, respectively. 
	$\lambda_n$ and $\lambda_c$ are set to 0.1 and 0.3 respectively, and $k$ is set to 10. $T_\text{warmup}$ is set to 50 epochs.
	For the classification objective, we initialize $\tau_t$ to 0.07 and set $\tau_s$ to 0.1, then warm up to 0.04 over the first 30 epochs using a cosine schedule. \\
	\noindent\textbf{Compared methods.}
	The compared methods include three competing novel category discovery methods RS+~\cite{han2021autonovel}, UNO+ \cite{fini2021unified} and k-means~\cite{macqueen1967some_kmeans} with DINO~\cite{caron2021} features, and five state-of-the-art GCD methods ORCA~\cite{cao2021orca}, GCD~\cite{vaze2022generalized}, XCon~\cite{fei2022xcon}, PromptCAL~\cite{zhang2023promptcal} and DCCL~\cite{pu2023dynamic}.

	\subsection{Experimental Results}
	
	We report the results of different methods in Tables~\ref{tab:ssb}-~\ref{tab:challenging}. In general, our method achieves the best performance in terms of ACC in most cases, which demonstrates the effectiveness of our model for GCD task. It is observed that compared with the novel category discovery methods RS+~\cite{han2021autonovel}, UNO+ \cite{fini2021unified} and k-means~\cite{macqueen1967some_kmeans} with DINO~\cite{caron2021} features, all six GCD methods yield consistently better results by a large margin, especially for the novel categories. Among GCD methods, the proposed method often performs better in `All', `Old' and `New' categories. Compared with SimGCD, our model consistently yields improved performance for all categories by 4.2\% in CUB, 2\% in Standford Cars, 2.6\% in FGVC-Aircraft, 1.9\% in CIFAR100 and 1.6\% in Image-Net-100. The performance improvement can mainly be attributed to the introduced contextual information, which facilitates the learned features to be more separable for the parametric classifier. Table~\ref{tab:challenging} shows the results on more challenging data sets, i.e., the imbalanced Herbarium-19 and the large-scale ImageNet-1k. All three GCD methods obtain remarkable performance improvement over the novel category discovery methods in Herbarium, which demonstrates the robustness of the GCD methods in real applications.

	
	\begin{table}[t]
		\caption{Results on more challenging datasets in terms of ACC}
		\resizebox{\linewidth}{!}{
			\begin{tabular}{@{}ccccccc@{}}
				\toprule
				& \multicolumn{3}{c}{Herbarium 19} & \multicolumn{3}{c}{ImageNet-1K} \\ \cmidrule(l){2-4} \cmidrule(l){5-7} 
				Methods       & All & Old & New  & All  & Old & New \\ 
				\midrule
				$k$-means~\cite{macqueen1967some_kmeans}  & 13.0 & 12.2 & 13.4 & - & - & - \\
				RS+~\cite{han2021autonovel}      & 27.9 & {55.8} & 12.8 & - & - & -  \\
				UNO+~\cite{fini2021unified}    & 28.3 & {53.7} & 14.7 & - & - & -  \\
				ORCA~\cite{cao2021orca}          & 20.9 & 30.9 & 15.5 & - & - & -  \\
				\midrule
				GCD~\cite{vaze2022generalized}   & {35.4} & 51.0 & {27.0} & 52.5 & 72.5 & 42.2  \\
				SimGCD~\cite{wen2023parametric}     & \ul{44.0} & \ul{58.0} & \ul{36.4} & \ul{57.1} & \ul{77.3} & \ul{46.9} \\  
				Ours                      & \textbf{45.7} & \textbf{58.9} & \textbf{38.3} & \textbf{59.5} & \textbf{79.9} & \textbf{49.5} \\
				\bottomrule
			\end{tabular}
		}
		\label{tab:challenging}
		\vspace{-.5cm}
	\end{table}

	\subsection{Ablation Study}
	To  verify the effectiveness of the contextual components in our model, we report in Table \ref{tab:losses} the results of the reduced versions of our model by removing the losses $\mathcal{L}_c$ and $\mathcal{L}_n$, respectively. The experiments are conducted on the datasets CUB and ImageNet-100. The results show that all three models incorporating contextual information yield improved performance compared with SimGCD in `All', `Old' and `New' categories, which demonstrates the advantages of our design. It is observed that the instance-level contextual constraint $\mathcal{L}_n$ makes a larger contribution over the cluster-level contextual constraint and combining the two contextual constraints leads a further improved performance.

	\begin{table}[h]
		\centering
		\caption{Ablation study}
		\label{tab:losses}
		\begin{tabular}{l ccc ccc}
			\toprule
			& \multicolumn{3}{c}{CUB}  & \multicolumn{3}{c}{ImageNet-100}\\
   \cmidrule(l){2-4} \cmidrule(l){5-7}
			Methods   &  All & Old & New & All & Old & New \\
			\midrule
			SimGCD   & 60.3 & 65.6 & 57.7 &  83.0 & 93.1 & 77.9         \\
			\midrule
			Ours w/o $\mathcal{L}_{n}$ & 61.4 & 66.5  & 58.2 & 83.5 & 93.4 & 79.1\\
			Ours w/o $\mathcal{L}_{c}$ & 62.0 & 66.0  & 59.0 & 84.1 & 93.7 & 80.2\\
			\midrule
			Ours & \textbf{64.5} & \textbf{68.7} & \textbf{61.5}  & \textbf{84.6} & \textbf{94.0} & \textbf{81.0}\\
			\bottomrule
		\end{tabular}
	\end{table}

	\subsection{The Influence of the Hyperparameters}
	To investigate the influence of hyperparameters on the performance of our model, we report the results of our model in Table \ref{tab:sensitivity} on the data sets CUB and ImageNet-100 by tuning $\lambda_n$, $\lambda_c$ and $k$ with different values. In general, our model is stable within certain ranges with respect to the different parameters. It is noticed that the performance of our model is degraded when over strong penalties are imposed on the contextual losses. In addition, a larger $k$ leads to a reduced performance, which is caused by the obtained inaccurate contextually similar pairs.
	\begin{table}[h]
		\centering
		\caption{The influence of the hyperparameters}
		\label{tab:sensitivity}
		\begin{tabular}{l ccc ccc}
			\toprule
			& \multicolumn{3}{c}{CUB}  & \multicolumn{3}{c}{ImageNet-100}\\
   \cmidrule(l){2-4} \cmidrule(l){5-7}
			Methods   &  All & Old & New & All & Old & New \\
			\midrule
			$\lambda_{n}=0.1$ & 64.5 & {68.7} & {61.5}  & {84.6} & {94.0} & {81.0}\\
			$\lambda_{n}=0.3$ & 64.7 & 68.9   & 61.6    &  84.3  &  93.6  &  80.6\\
			$\lambda_{n}=0.5$ & 63.5 & 67.3   & 60.7    &  83.5  &  93.0  &  79.8\\
			$\lambda_{n}=0.7$ & 61.2 & 64.5   & 58.9    &  81.4  & 89.6   &  78.6\\
			\midrule
			$\lambda_{c}=0.1$ & 63.8 & 67.1 & 60.1 &      83.4   &   93.2   & 80.0\\
			$\lambda_{c}=0.3$ & {64.5} & {68.7} & {61.5}  & {84.6} & {94.0} & {81.0}\\
			$\lambda_{c}=0.5$ & 63.7   &  67.9  &  59.8  &  83.0  & 93.1  &  79.8\\
			$\lambda_{c}=0.7$ &  61.5  &  66.5  &  58.7  &  81.4  &  92.3  &  78.5\\
			\midrule
			$k=10$ & {64.5} & {68.7} & {61.5}  & {84.6} & {94.0} & {81.0}\\
			$k=20$ & 63.4  &  67.8  &  60.8    &  83.5  & 92.4  &  80.2\\
			$k=30$ & 62.8 & 66.8   &  59.7   &  82.6  &  91.5  &  79.5\\
			\bottomrule
		\end{tabular}
	\end{table}



	\section{Conclusion}
	In this study, we propose a novel semi-supervised approach to address the challenging GCD problem, which mainly consists of a  feature representation module and a classifier learning module. Particularly, inspired by human cognition, we incorporate contextuality into the representation learning to make the learned feature more discriminative for both seen and novel categories. This results in two novel contextual contrastive learning losses at instance-level and cluster-level, promoting to minimize the intra-class variance and maximize the inter-class distance. The improved features enhance the performance of the jointly learned parametric classifier, which in turns facilitates the feature learning.	Extensive experiments demonstrate the superior performance of our model over the state-of-the-art.

	{
		\small
		\bibliographystyle{IEEEtran}
		\bibliography{strings,refs}
	}
\end{document}